\documentclass{article}
\usepackage[preprint]{eiml_style_2025}

\usepackage{graphicx}
\usepackage{booktabs}
\usepackage{amsmath, amssymb}
\usepackage{hyperref}
\usepackage{multirow}
\usepackage{xcolor}
\title{Adaptive Individual Uncertainty under Out-Of-Distribution Shift with Expert-Routed Conformal Prediction}
\author{
  Amitesh Badkul \\
  Ph.D. Programs in Computer Science \\
  The Graduate Center, City University of New York \\
  \texttt{abadkul@gradcenter.cuny.edu}
  \And
  Lei Xie \\
  School of Pharmacy and Pharmaceutical Sciences \\ Center for Drug Discovery \\
  Northeastern University \\
  \texttt{lxie@iscb.org}
}

\begin{document}
\maketitle

\begin{abstract}
Reliable, informative, and individual uncertainty quantification (UQ) remains missing in current ML community. This hinders the effective application of AI/ML to risk-sensitive domains. Most methods either fail to provide coverage on new data, inflate intervals so broadly that they are not actionable, or assign uncertainties that do not track actual error,  especially under a distribution shift. In high-stakes drug discovery, protein–ligand affinity (PLI) prediction is especially challenging as assay noise is heterogeneous, chemical space is imbalanced and large, and practical evaluations routinely involve distribution shift. In this work, we introduce a novel uncertainty quantification method, \textbf{T}rustworthy \textbf{E}xpert \textbf{S}plit-conformal with \textbf{S}caled Estimation for \textbf{E}fficient \textbf{R}eliable \textbf{A}daptive intervals (TESSERA), that provides per-sample uncertainty with reliable coverage guarantee, informative and adaptive prediction interval widths that track the absolute error. We evaluate on protein–ligand binding affinity prediction under both independent and identically distributed (i.i.d.) and scaffold-based out-of-distribution (OOD) splits, comparing against strong UQ baselines. TESSERA attains near-nominal coverage and the best coverage–width trade-off as measured by the Coverage–Width Criterion (CWC), while maintaining competitive adaptivity (lowest Area Under the Sparsification Error (AUSE)). Size-Stratified Coverage (SSC) further confirms that intervals are right-sized, indicating width increases when data are scarce or noisy, and remain tight when predictions are reliable. By unifying Mixture of Expert (MoE) diversity with conformal calibration, TESSERA delivers trustworthy, tight, and adaptive uncertainties that are well-suited to selective prediction and downstream decision-making in the drug-discovery pipeline and other applications.
\end{abstract}

\section{Introduction}
Modern Artificial Intelligence (AI) systems are increasingly deployed inside risk-sensitive pipelines where mistakes carry high costs in human safety, time, and money. In autonomous driving, perception–planning–control components must make split-second decisions from uncertain, partial sensory input. The deep learning (DL) models mapping pixels to operate the vehicle emphasize the need for reliable confidence information to decide when to slow down, hand control over to a human, or re-plan under ambiguous conditions \cite{bojarski2016end}. Similarly, in medical imaging-based diagnostics, deep networks can match specialist performance for certain tasks, but their utility in practice depends on knowing when not to trust a prediction \cite{esteva2017dermatologist}. The same precedent holds for machine learning (ML) models usage in various stages of drug discovery including but not limited to virtual screening, lead optimization and hit triage. These examples necessitate that decisions should depend not just on accurate point estimates, but on individualized, trustworthy uncertainty that reflects case-by-case risk. Beyond these safety-critical examples, uncertainty quantification (UQ) has become standard practice in other AI areas. In computer vision, Bayesian deep learning work explicitly models aleatoric and epistemic uncertainty and shows task gains in semantic segmentation and depth regression, helping systems decide when predictions are unreliable \cite{kendall2017uncertaintiesneedbayesiandeep}. In natural language processing, multiple studies and surveys now evaluate calibration and uncertainty for pre-trained and large language models, using UQ for selective prediction and robustness under shift \cite{geng2023survey, xiao2022uncertainty}. By contrast, adoption in drug discovery remains more limited and uneven. Several works explain uncertainty for molecular property prediction tasks \cite{hirschfeld2020uncertainty}. Conformal Prediction is now becoming more prevalent, as a way to enforce finite-sample coverage, underscoring the need for reliability guarantees in this domain, but all these works have inconsistent reporting and calibration and often degrade under distribution shift. The drug-discovery setting is intrinsically challenging for UQ. First, the underlying chemical space is vastly large, consisting of over $10^{60}$ compounds \cite{reymond2015chemical}. Second, even major bioactivity resources annotate only on the order of $10^6$  compounds against a few thousand protein targets, leaving majority compounds under-represented \cite{gaulton2012chembl}. Third, practical evaluations routinely involve distribution shift, where models must generalize beyond i.i.d. assumptions. These facts motivate our focus on protein–ligand affinity (PLI) prediction as a representative, high-stakes use case. PLI couples heterogeneous assay noise and imbalanced as well as underrepresented scaffolds with real-world decisions about which candidates to advance to further stages in drug discovery. 

The modern uncertainty-quantification (UQ) toolbox spans both Bayesian and non-Bayesian estimators. On the Bayesian side, Monte Carlo Dropout \cite{gal2016dropout} and related approximations aim to capture epistemic uncertainty. The non-Bayesian side includes deep ensembles \cite{lakshminarayanan2017simplescalablepredictiveuncertainty}, Gaussian processes (GPs) \cite{seeger2004gaussian, qiu2019quantifying}, and distance-based methods \cite{badkul2024emosaic}. These methods are necessary because modern neural networks are systematically miscalibrated and often overconfident even when wrong \cite{guo2017calibration}, so raw scores cannot be trusted in risk-sensitive pipelines. However, large-scale evaluations show that, while many of these methods calibrate reasonably on i.i.d. data, their uncertainty quality degrades under distribution-shift/data-shift \cite{ovadia2019can} due to inherent assumptions associated with the distribution of the data, and the residuals as well as the method's properties, yielding narrow but wrong intervals. To mitigate this issue of distribution dependent modeling, researchers introduced called conformal prediction (CP) \cite{angelopoulos2021gentle}, a distribution-free wrapper for uncertainty, given any trained predictor, CP returns finite-sample, model-agnostic prediction sets/intervals that achieve target marginal coverage, a compelling property in case of safety-critical use. In regression, however, the vanilla split-CP recipe that calibrates a single residual scale typically yields constant or weakly varying widths, which can be inefficient in heteroscedastic settings, for the vast majority of practical applications. This inefficiency, and the need for adaptivity, is a central motivation behind Conformalized Quantile Regression (CQR). CQR combines CP with quantile regression so that widths track the conditional label distribution, while retaining distribution-free coverage. In practice, though, CQR introduces several limitations one must train multiple quantile heads with the pinball loss, and enforce non-crossing among quantiles, all of which add hyperparameters and failure modes to the pipeline. For additional background on modern UQ methods, MoE-based UQ, and UQ in drug-discovery we refer readers to the appendix section \ref{sec:related-works}.

Our approach TESSERA (\textbf{T}rustworthy \textbf{E}xpert \textbf{S}plit-conformal with \textbf{S}caled estimation for \textbf{E}fficient  \textbf{R}eliable \textbf{A}daptive intervals), takes these limitations into account and builds a model-aware difficulty scale directly from a Mixture-of-Experts (MoE) backbone, alongside an explicit decomposition of uncertainty into epistemic and aleatoric components. Expert disagreement quantifies lack of consensus among routed experts or the epistemic uncertainty, and per-expert variance heads capture prediction spread at the expert level or the Aleatoric uncertainty. We then apply a single split-conformal calibration to this scale to produce per-sample prediction intervals with finite-sample, distribution-free coverage. Intervals that are widen where the model is uncertain and remain tight where experts agree. The entire pipeline including the architecture of the model as well as detailed description of the dataset, baselines, and the evaluation metrics are presented in the appendix section \ref{sec:method}. The hyperparameters associated with different models and method are described in \ref{sec:hyperparams}\ref{tab:tessera-hyperparam} and \ref{tab:baseline-hyperparam} for the TESSERA and baseline methods. Throughout the work, we denote TESSERA for CP-calibrated variants and adopt the following shorthand: TESSERA$_A$, and TESSERA$_E$ denote conformalized intervals built from the expert-disagreement (epistemic) and per-expert variance (aleatoric) signals respectively. The MoE$_E$ and MoE$_A$ refer to the corresponding raw (uncalibrated) MoE proxies. For additional analyses and supporting results, readers are directed to Appendix Section \ref{sec:disent}.

Our contributions include:
\begin{enumerate}\vspace{-2mm}
    \item Reliable, informative, and adaptive uncertainty. TESSERA delivers nominal coverage (validity), competitive widths (efficiency), and strong adaptivity (intervals track local difficulty), by calibrating a model-aware scale rather than inflating widths globally.
    \item Robustness under distribution shift. The MoE-decomposed uncertainty heuristics concentrates disagreement in OOD regions, and split conformal maps it to coverage-guaranteed intervals, thereby making TESSERA suitable for OOD shift.
\end{enumerate}

\section{Results and Discussion}

\begin{table}[h]
\centering
\caption{\textbf{Comparison of uncertainty–quality metrics.} Test–set results for baselines (MC Dropout, Classical CP, RIO-GP, eMOSAIC), MoE$_E$, MoE$_A$, and TESSERA$_{E}$, and TESSERA$_{A}$ with $\alpha = 0.10$. 
PICP targets $1-\alpha \approx 0.90$ (higher is better). MPIW and NMPIW assess interval efficiency (lower is better). AUSE evaluates the quality of the uncertainty ranking via sparsification error (lower is better). CWC is calculated with penalty $\eta{=}50$ (lower is better). Best values are \textbf{bold}, and the second–best are \underline{underlined}.}
\label{tab:uncertainty-metrics}
\begin{tabular}{@{}clccccc@{}}
\toprule
\multicolumn{2}{c}{Method} & PICP ($\uparrow\;{\approx}0.9$) & MPIW ($\downarrow$) & NMPIW ($\downarrow$) & AUSE ($\downarrow$) & CWC ($\downarrow$) \\
\midrule
\multirow{4}{*}{Baselines}
  & Monte Carlo Dropout & 0.16 & \textbf{0.37} & \textbf{0.02} & \underline{0.59} & 25.23 \\
  & RIO-GP              & 0.27 & \underline{0.71} & \underline{0.03} & 0.74 & 15.70 \\
  & Classical CP             & 0.91 & 3.97 & 0.17 & 0.80 & 0.17 \\
  & eMOSAIC             & \underline{0.64} & 2.01 & 0.08 & 0.74 & 1.22 \\
\midrule
\multirow{4}{*}{Ours}
  & MoE$_E$ & 0.48 & 1.49 & 0.06 & \textbf{0.58} & 4.12 \\
  & MoE$_A$  & 0.40 & 1.07 & 0.04 & 0.64 & 6.98 \\
  & TESSERA$_{E}$ & \textbf{0.91} & 4.03 & 0.17 & 0.64 & \textbf{0.17} \\
  & TESSERA$_{A}$  & \textbf{0.91} & 4.76 & 0.20 & \textbf{0.58} & \underline{0.20} \\
\bottomrule
\end{tabular}
\end{table}

\subsection{Valid Coverage at the nominal level across Out-of-Distribution Scaffolds}
\begin{figure}[h]
    \centering
    \includegraphics[width=1\linewidth]{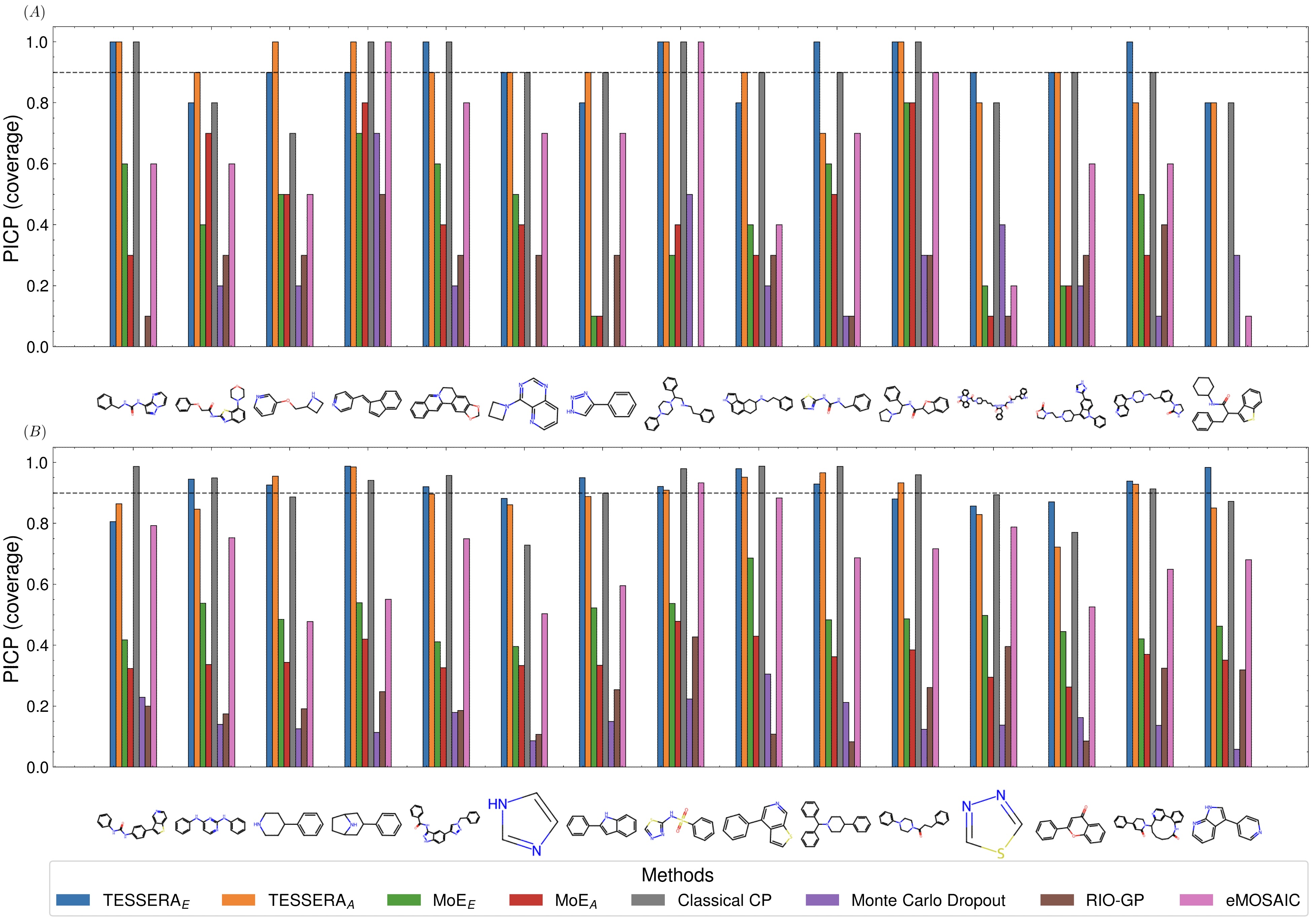}
    \caption{Scaffold-wise coverage under distribution shift. (A) Prediction-interval coverage probability (PICP) computed per Bemis–Murcko scaffold for the 15 most frequently occuring scaffolds in the test set. (B) The same analysis for the 15 least frequent scaffolds. The dashed line marks the nominal target $1-\alpha$=0.90. Bars are color-coded by methods and the scaffold sketches below the axis correspond to the x-axis groups. The panels highlight how coverage varies across scaffolds and under rarity and shift. TESSERA generally remain closer to the target line, whereas non-calibrated baselines tend to under-cover on rare scaffolds.}
    \label{fig:scaffold-shift}
\end{figure}

We compare TESSERA to baselines (MC Dropout, RIO-GP, eMOSAIC, Classical CP) and to MoE-based raw heuristics (expert disagreement, predicted variance). On the scaffold–OOD split, TESSERA$_A$, TESSERA$_E$, and classical CP attain near-nominal coverage ($PICP\approx0.91$ at $1-\alpha= 0.90$), while the baselines systematically under-cover Table \ref{tab:uncertainty-metrics}. We can clearly see the strength of TESSERA, which on average is better $83.8\%$ over the baselines, and performs at least $90.9\%$ over all the models being compared. Figure \ref{fig:scaffold-shift} further shows that CP methods remain close to the nominal level per scaffold family in the test set, both for the 15 most frequent and the 15 least frequent scaffolds (minimum $n\geq10$ compounds), whereas non-CP methods vary widely and often miss the target on rare scaffolds. MC Dropout interprets dropout as approximate Bayesian inference \cite{gal2016dropout}, but its uncertainty quality is sensitive to the dropout rate and the coarseness of the variational approximation. Large empirical studies show that MC Dropout degrades under distribution shift and is often miscalibrated \cite{ovadia2019can}. In regression, additional analyses find that MC Dropout uncertainties are prone to miscalibration and often need post-hoc scaling to restore calibration \cite{laves2020calibration}. These behaviors match Fig. \ref{fig:uq_overview}(C), where narrow bands lead to systematic under-coverage. While RIO can improve means and yield residual variances when its assumptions hold \cite{qiu2019quantifying}, two properties of our setting explain the observed under-coverage on OOD scaffolds. Standard residual GPs assume homoscedastic Gaussian noise, our residuals are heavy-tailed and heteroscedastic, so posterior variances are misscaled unless one explicitly models input-dependent noise \cite{lazaro2011variational}. Moreover, the residual GP receives is trained on roughly $210{,}240$ data points, where distance-based kernels suffer from distance concentration and over-smoothing, harming extrapolation and calibration \cite{binois2022survey,capone2022gaussian}. These limitations align with our experimental evidence in Fig.\ref{fig:scaffold-shift}, where RIO–GP fails on several rare scaffolds, yielding intervals that are narrow yet systematically under-cover. eMOSAIC provides a strong OOD ranking signal via latent-space clustering and Mahalanobis distance utilization \cite{badkul2024emosaic}, which explains why it beats raw MoE heuristics. However, in high dimensions, covariance estimation and distance concentration further compress anomaly scores, yielding narrow but under-covering intervals\cite{aggarwal2001surprising, chen2011robust}. Classical CP attains the target marginal coverage by design, so its PICP is expected to match ours. Comparison of the MoE-based raw heuristics with TESSERA-based variants on i.i.d. split are available in the appendix section \ref{sec:ablation}, and in Table \ref{sec:ablation}\ref{tab:uncertainty-split-comparison} 

\subsection{Efficient and Informative narrow intervals}
\begin{figure}[t]
    \centering
    \includegraphics[width=\linewidth]{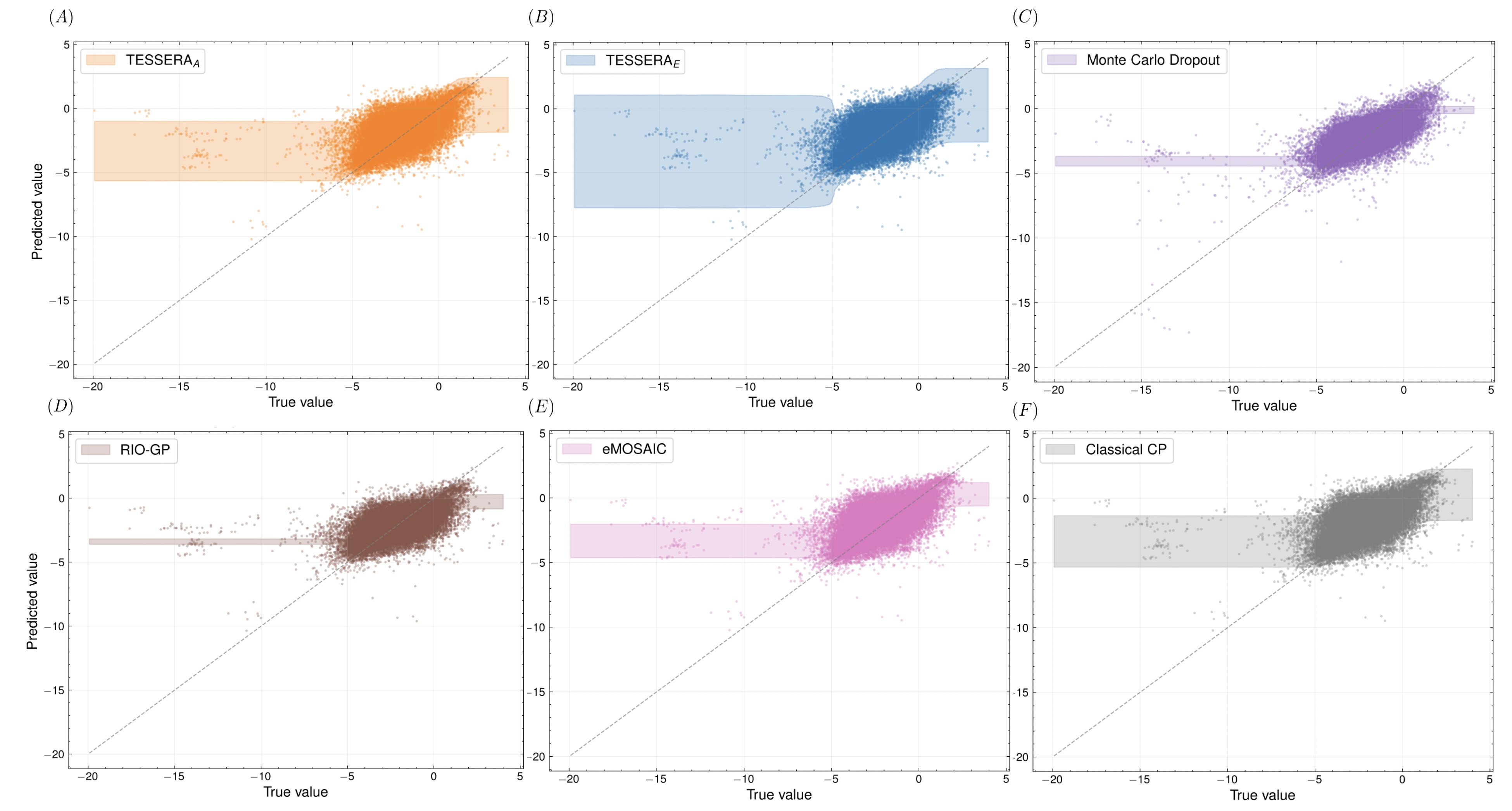}
    \caption{\textbf{Visualizing predictive intervals and adaptivity across methods.}
    (A–F) Scatter of predictions $\hat{y}$ versus ground truth $y$ on the test set. 
    The semi-transparent band show the method’s prediction intervals (smoothed along sorted $y$ for visual clarity), and the dashed line is $y{=}x$. 
    Panels correspond to: (A) CP–Predicted Variance, (B) CP–Expert Disagreement, (C) Monte Carlo Dropout, (D) RIO-GP, (E) eMOSAIC, and
    (F) Classical CP.}
    \label{fig:uq_overview}
\end{figure}

One of our other goals is to keep intervals narrow while maintaining nominal coverage. We therefore report both width-only metrics including MPIW, and NMPIW, along with the combined CWC, which penalizes under-coverage and otherwise reduces to normalized width. On the scaffold–OOD split, the Table \ref{tab:uncertainty-metrics} reflects that the conformal variants are not only valid but also efficient. While the best and second best MPIW, and NMPIW, as indicated from the Table \ref{tab:uncertainty-metrics}, are the MC Dropout and RIO-GP. However, we can clearly see that they achieve the worst coverage, and highest under-coverage. Essentially, they have narrow intervals all across the test set from Fig. \ref{fig:uq_overview}(C)(D), making their uncertainty uninformative. TESSERA$_A$ and TESSERA$_E$ achieve the best CWC, along with classical CP, on average $98.3\%$ better performance when compared with baselines and MoE-based heuristics, the same is reflected across the Fig. \ref{fig:uq_overview}(A)(B). We include a sensitivity Table \ref{sec:ablation}\ref{tab:cwc-eta} for $\eta \in {10, 50, 100}$ and observe unchanged method ranking. eMOSAIC as discussed, doesn't have severe under-coverage as well as not too wide or narrow intervals, therefore having the best performance among baselines even beating our raw uncertainty heuristics from MoE by substantial margin. 

MC Dropout correlates to a mean-field variational approximation, and mean-field bayesian neural networks (BNNs) are known to underestimate posterior variance and push the predictive variance downward and making intervals small. Additionally, MC Dropout's confidence on shifted inputs or OOD space remains high, this manifests as narrow prediction intervals \cite{ovadia2019can}. For RIO-GP, computing variance at fixed, learned hyperparameters may makes the GP error bars a lower bound on prediction error, and further tightening bands \cite{wagberg2017prediction}. For eMOSAIC, mapping latent distance scores to widths without robust calibration leads to compressed prediction intervals.  Along with this feature-norm and layer sensitivity for selection of the modeling, further reduce the prediction intervals \cite{mueller2025mahalanobis++}. Classical CP calibrates a single residual scale, producing intervals of nearly constant width across inputs. In real-world with heteroscedasticity settings this is inefficient, intervals remain wide even where predictions are easy and too narrow where noise is high. This well-known inefficiency is precisely why conformalized quantile regression was introduced to obtain input-dependent widths \cite{romano2019conformalized}.  

\subsection{Adaptive and Actionable Uncertainties}
\begin{figure}
    \centering
    \includegraphics[width=1\linewidth]{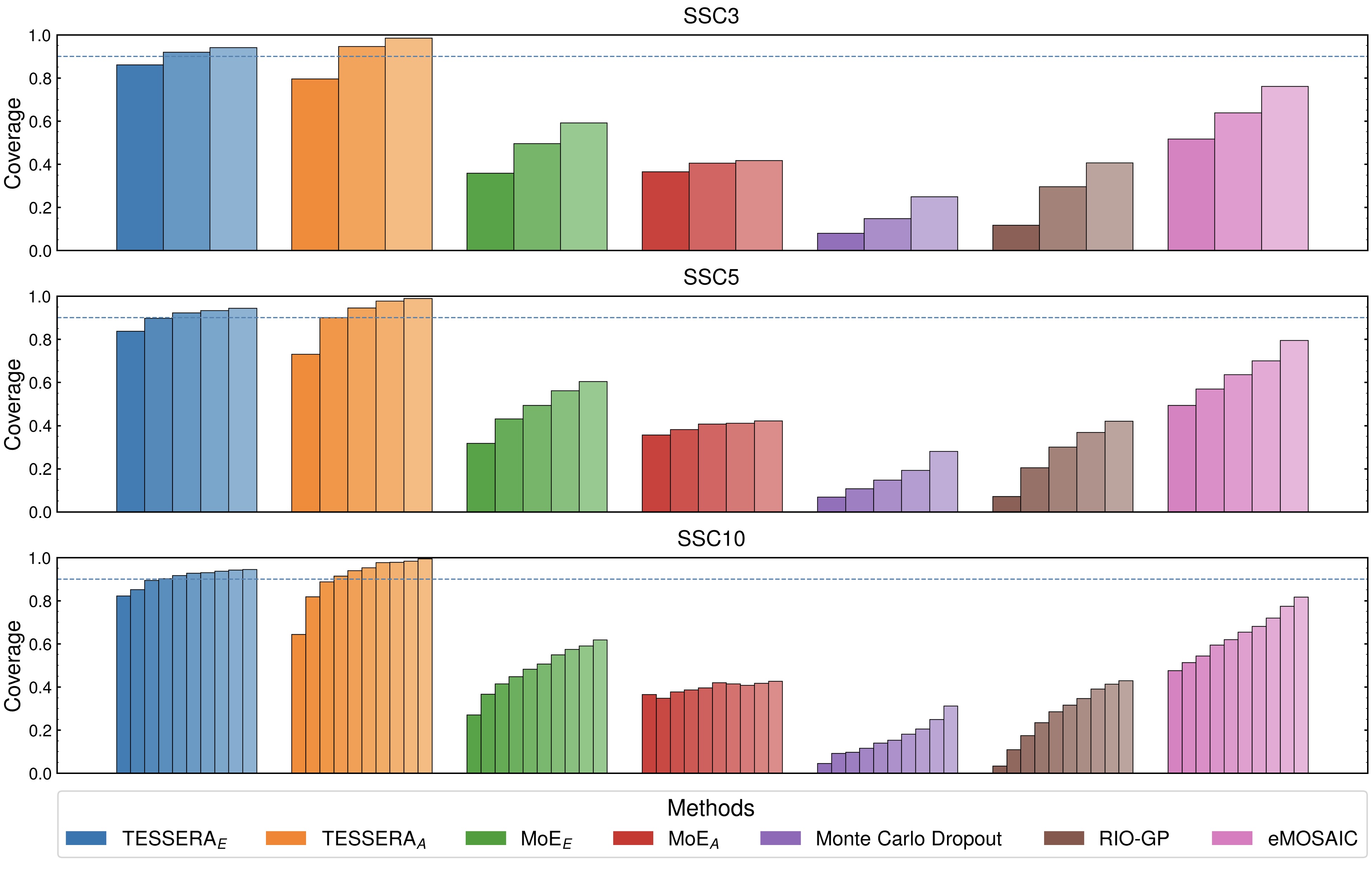}
    \caption{\textbf{Size–Stratified Coverage (SSC).} For each method, test points are sorted by prediction–interval (PI) width and split into equal–sized bins (left\,$\rightarrow$\,right: narrow\,$\rightarrow$\,wide). Bars show the empirical coverage within each bin. The three panels report SSC with $J\!\in\!\{3,5,10\}$ bins (top to bottom). The dashed horizontal line indicates the nominal target coverage $1-\alpha$ (here, $0.90$). Ideally, all bars should lie near the dashed line, indicating conditionally well–calibrated PIs across easy (narrow) and hard (wide) regions. We compare our method with various baselines. We omit SSC for Classical CP because its split-residual construction produces constant-width intervals making SSC uninformative, SSC is designed to assess adaptive methods whose widths vary with difficulty.}
    \label{fig:ssc_coverage}
\end{figure}

Adaptivity asks whether a method’s uncertainty orders test points by difficulty: as we remove the fraction of points with the largest predicted uncertainty, does the error on the remaining set decrease quickly? To check that ordering translates into trustworthy decisions, we report the SSC, we bin test examples by their returned PI width (narrow to wide) and compute empirical coverage in each bin. The other baseline methods climb from large under-coverage in the narrow bins to better coverage only in the widest bins, indicating that their ordering has some uncertainty signal, but their absolute scale is too small for validity, again reinforcing results of the validity and efficiency metrics regarding their prediction interval widths and coverage. Our TESSERA model's coverage across all bins as seen in Fig. \ref{fig:ssc_coverage}. Along with this, from Fig. \ref{fig:ause_plot}, our TESSERA$_E$ variant yields the lowest AUSE among all methods, indicating that its uncertainty most closely tracks true error, and TESSERA$_A$ is competitive. By contrast, eMOSAIC and RIO–GP achieve higher AUSE, reflecting weaker alignment between their scores and realized errors. In Fig. \ref{fig:ssc_coverage}, results from TESSERA remain near the 0.90 dashed line within each width bin, from narrow to wide, demonstrating that the intervals are "correct" even on easy cases. 
Uncertainty-guided acquisition has repeatedly improved label efficiency in practice, from image classification and medical imaging \cite{gal2017deep, kirsch2019batchbald} to materials and drug discovery \cite{yin2023evaluating, tosh2025bayesian}. In our setting, CP-based widths exhibit the two properties and prerequisites that these systems rely on, the strong error ranking and reliable per-bin coverage. We therefore hypothesize that selecting molecules by CP interval width would yield actionable acquisitions, improving virtual screening efficiency. Classical residual CP applies a single residual quantile to all inputs, so interval widths are (nearly) constant and do not adapt to local difficulty, this yields valid marginal coverage but poor or no informativeness. 

\section{Conclusion}
TESSERA learns two model-aware difficulty signal with a Mixture-of-Experts backbone and mapping it to per-sample intervals with a single split-conformal calibration yields uncertainties that are valid, tight, and adaptive in practice. On protein–ligand affinity prediction, we observed near-nominal coverage across both scaffold (OOD) and random (i.i.d.) splits, competitive interval efficiency, strong sparsification behavior, and consistent results across three very different chemical encoders. Together, these findings indicate that MoE diversity provides a useful local scale for uncertainty while conformal calibration supplies the finite-sample coverage guarantee. However, there exists certain limitations and questions to be addressed. First, our analysis of the proposed epistemic (expert disagreement) and aleatoric (per-expert variance) components is intentionally conservative: because ground-truth decompositions are unobservable, disentanglement can only be assessed indirectly. A stronger validation needs synthetic controls and targeted perturbations. Second, our OOD evaluation focuses on ligand-side scaffold shift. In drug discovery, other shifts matter: protein splits probe generalization across unseen targets, we plan to test TESSERA under such protein-conditioned splits next. Lastly, while conformal prediction guarantees finite-sample marginal coverage under exchangeability, these guarantees can degrade when the test distribution differs from calibration and distribution-free conditional guarantees are known to be impossible in general. This motivates shift-aware conformal variants including but not limited to weighted CP under covariate shift, as well as  adaptive conformal inference as natural next steps for TESSERA. 

\bibliographystyle{plain}
\bibliography{ref}
\newpage
\appendix
\section{Related Work}\label{sec:related-works}
\subsection{Uncertainty Quantification (UQ)}
Early variational Bayesian neural networks attempted to place distributions over weights to represent parameter uncertainty. Graves introduced a practical stochastic variational method for neural networks \cite{graves2011practical}, reformulating common regularizers through a variational view and making Bayesian treatments practical. However, tractability typically required factorized or local Gaussian posteriors, which miss multi-modality and between-mode uncertainty. Bayes by Backprop \cite{blundell2015weight} then formalized a backpropogation-compatible variational objective to learn weight distributions end-to-end. It is simple to implement, and yielded stable predictive uncertainty but it still underestimates variance and its robustness degrades under data shift.  To reduce the computational burden of full Bayesian inference, Gal et al. \cite{gal2016dropout} introduced MC Dropout, that kept the dropout active at inference time as an approximate bayesian inference method, this method required almost no overhead computation as compared to the previous ones, however, studies showed that under OOD shift, the quality fo uncertainty degraded. Lakshminarayanan et al. \cite{lakshminarayanan2017simplescalablepredictiveuncertainty} offered a complementary non-Bayesian route is Deep Ensembles, which train multiple independently initialized networks to learn the distribution of uncertainty, leading to higher uncertainty in OOD setting as well as improved error uncertainty correlation. Another practical family uses Gaussian processes (GPs). Classical GPs place a distribution over functions and provide closed-form predictive means and variances, yielding coherent uncertainty estimates that have been widely regarded as well-calibrated, especially in small- to medium-size datasets. By providing an explicit predictive distribution, GPs capture a spectrum of likely outcomes and their probabilities \cite{seeger2004gaussian}.  Building on this idea, Qiu et al. \cite{qiu2019quantifying} proposed RIO which fits a GP post-hoc model agnostic method on the trained neural network's residuals over the input space improving both the means and uncertainty. Performance, however, depends on kernel assumptions and hyperparameters, and their reliability weakens under distribution shift. A different line estimates uncertainty from distance in representation space. eMOSAIC \cite{badkul2024emosaic} adopts latent clustering and Mahalanobis distance calcalculation as part of their pipeline, practical gains but being sensitive to the chosen feature layer, covariance conditioning and high-dimensional distance concentration collapse. Finally, conformal prediction (CP) \cite{angelopoulos2021gentle} provides a model-agnostic, distribution-free wrapper that yields finite-sample marginal coverage when placed atop any predictor. Classical residual CP can be ineffective due to constant prediction interval widths, motivating the development of CQR \cite{romano2019conformalized} which conformalizes learned conditional quantiles to obtain the per-sample adaptive widths.  However, the training of these CQR base models comes at a cost of optimizing the quantile-head training with pinball-loss and putting non-crossing constraints. 

\subsection{Mixture-Of-Experts as an Uncertainty Quantification Tool}
Researchers have also explored Mixture-of-Experts architectures specifically as a tool for uncertainty quantification. The idea is that different experts can capture different hypotheses through various signals and can serve as an indicator of uncertainty. Choi et al. \cite{choi2018uncertainty} introduced the idea of utilizing Mixture Density Network (MDN) to get sampling-free predictive mean and variance. Additionally, the authors explicitly decompose variance into between-component vs within-component variance. They show the decomposition distinguishes absence of data, heavy noise, and composition of functions, this abstraction indicates the separation of aleatoric and epistemic components of uncertainty. The MDN is a predecessor and subset of the modern Mixture-of-Experts (MoE). MoEs route the input through sub-networks each of which produces a point prediction, whereas an MDN routes only the output density parameters. Hence, MDNs can be viewed as output-space MoEs. Gao et al. \cite{gao2022modeling} introduce a Mixture of Stochastic Experts (MoSE) model for image segmentation to model multimodal aleatoric uncertainty. In segmentation tasks, there might be numerous ways to label an image, and MoSE explicitly accounts for this by letting each expert output a distinct segmentation, yielding better uncertainty characterization on certain ambiguous images than a single-model approach. However, they only model the aleatoric uncertainty, and the work doesn't conduct a reliability or coverage evaluation.  
Pavlitska et al. \cite{pavlitska2025extracting} recently present a principled framework for extracting uncertainty estimates from MoE models in semantic segmentation. They propose extraction of uncertainty from the MoE without any architectural changes to it including the following uncertainty heuristics - predictive entropy, mutual information, and expert variance. The authors empirically demonstrate that uncertainty signals extracted from the MoE tend to be more reliably calibrated under distributional shift than those from conventional deep ensembles. However, raw uncertainty scores from MoEs are not guaranteed to be probabilistically calibrated. The combination of MoEs with conformal prediction is an exciting avenue, as it could merge learned uncertainty estimates with formal coverage control – ensuring that the improved diversity MoEs offer actually translates into trustworthy confidence intervals in practice. 

\section{Method}\label{sec:method}
Our TESSERA model consists of two major components, the first component is a MoE which predicts the mean and variance through each of it's experts, by minimizing the negative log likelihood (NLL) loss. From the various predictions of the experts we obtain the aleatoric, and epistemic uncertainty through the different means and variances. Once we have these raw heuristics, we use the second major component of our pipeline, which is performing the split conformal prediction using our  per-sample raw uncertainty heuristic as part of the nonconformity score $\alpha_i$ on the calibration set which is then used to obtain the personalized intervals for the unknown points. 

\paragraph{MoE as UQ}
We use protein-ligand binding affinity prediction, an important but challenging problem \cite{wang2021computationally}, as a case study. We model the protein–ligand binding affinity prediction as a supervised regression problem $y=f(z_{\text{pli}})$ where $z_{\text{pli}}$ is a fused representation of a protein sequence and a small molecule (SMILES). The protein sequence is embedded by a protein encoder $z_p$, while the molecule is embedded by a chemical encoder, using $z_c$. We concatenate the two modality embeddings:
\begin{equation}
z_{\text{pli}} = z_p \oplus z_c 
\end{equation}

We provide this $z_{\text{pli}}$ as input to the MoE, specifically a dense MoE, with $K$ experts 
produces expert-specific means and variances 
$\{\mu_k(z_{\text{pli}}),\sigma_k^2(z_{\text{pli}})\}_{k=1}^K$ 
by using a gating mechanism with gating weights $w_k(z_{\text{pli}})$ obtained by passing the $z_{\text{pli}}$ through a non-linear multi-layer perceptron (MLP) with softmax such that $\sum_k w_k(z_{\text{pli}})=1$. 

Given the MoE with input $z_{\text{pli}}$ the conditional predictive density is a mixture of Gaussians:
\begin{equation}
    p(y\mid z_{\text{pli}})=\sum_{k=1}^{K} 
w_k(z_{\text{pli}})\,\mathcal{N}\!\big(y;\mu_k(z_{\text{pli}}),\sigma_k^2(z_{\text{pli}})\big)
\end{equation}
with predictive mean,
\begin{equation}
    \hat\mu(z_{\text{pli}})=\sum_k w_k(z_{\text{pli}})\,\mu_k(z_{\text{pli}})
\end{equation}
MoEs act as structured, parameter-sharing approximations to deep ensembles, offering ensemble-like diversity at lower inference cost via learned routing. We train the MoE with Gaussian NLL so that each expert’s variance head learns input-dependent noise:
\begin{equation}
\mathcal{L}_{\mathrm{mixNLL}} = -\frac{1}{n}\sum_{i=1}^{n}  \log\!\left(\sum_{k=1}^{K}w_k(z^{(i)}_{\mathrm{pli}})\, \frac{1}{\sqrt{2\pi\sigma_k^2(z^{(i)}_{\mathrm{pli}})}}\,
\exp\!\Big[-\tfrac{(y_i-\mu_k(z^{(i)}_{\mathrm{pli}}))^2}{2\sigma_k^2(z^{(i)}_{\mathrm{pli}})}\Big]\right)
\end{equation}
NLL is a proper scoring rule for probabilistic regression and is standard when learning aleatoric uncertainty \cite{kendall2017uncertaintiesneedbayesiandeep}.

In TESSERA, the \emph{aleatoric} and \emph{epistemic} uncertainties are decomposed for any input $x=z_{\mathrm{pli}}$ as follows: 

\begin{enumerate}
    \item \textbf{Aleatoric} or data-associated uncertainty which we obtain from expert variance heads
    \begin{equation}
        A(x) = \sqrt{\sum_{k=1}^{K} w_k(x)\,\sigma_k^2(x)} .
    \end{equation}
    This reflects the level of observation noise the model has learned to attribute to the input. Training with NLL makes this signal meaningful as input-dependent noise \cite{kendall2017uncertaintiesneedbayesiandeep}. Since we obtain this heuristic from the head of the neural network model, we call this as predicted variance. 
    \item \textbf{Epistemic} or model-associated uncertainty, obtained from expert disagreement in predicted means, because gating can be peaky, we use the unweighted deviation as the heuristic. We call this as expert disagreement, it is a well-used epistemic proxy in ensemble-like systems including deep ensembles \cite{lakshminarayanan2017simplescalablepredictiveuncertainty} and MoE \cite{pavlitska2025extracting}, across numerous domains \cite{gawlikowski2023survey}. 
    \begin{equation}
    E(x) = \sqrt{\frac{1}{K}\sum_{k=1}^{K}\big(\mu_k(x)-\bar\mu_K(x)\big)^2},
    \qquad
    \bar\mu_K(x)=\frac{1}{K}\sum_{k=1}^{K}\mu_k(x)
    \end{equation}

\end{enumerate}

\paragraph{Individualized Split Conformal Prediction} 
We adopt a split conformal regression framework with a normalized nonconformity score. After training our MoE model on a training set, we calibrate the intervals on a held-out calibration set $\{(x_i,y_i)\} \in C$, by computing the nonconformity score: 
\begin{equation}
\tilde\alpha_i = \frac{|y_i-\hat\mu(x_i)|}{S(x_i)+\varepsilon},
\end{equation}
where $S(\cdot)\in\{E,\,A\}$, essentially we use the raw uncertainty heuristics obtained from the decomposition of MoE means and variances. We define $\hat q_{1-\alpha}$ to be the $(1-\alpha)(1+1/n_{\text{cal}})$ quantile of $\{\tilde{\alpha}_1,\dots,\tilde{\alpha}_{n_{\text{cal}}}\}$, where $n_{\text{cal}}$ is the number of calibration samples. The extra $(1+1/n_{\text{cal}})$ factor is a standard finite-sample adjustment for split CP. Intuitively, if $S(x_i)$ accurately reflects the scale of the error $|y_i - \hat{\mu}(x_i)|$, then the nonconformity score $\tilde{\alpha}_i$ should be roughly exchangeable across the calibration set.

For any test input $x$, the individualized interval is
\begin{equation}
\mathcal{I}_{1-\alpha}(x) = \hat\mu(x)\pm \hat q_{1-\alpha}\,S(x).
\end{equation}

We compute split-conformal prediction sets/intervals on a held-out calibration set. The standard finite-sample coverage guarantee for split conformal requires exchangeability between the calibration and test examples (a condition slightly weaker than i.i.d.). Because our scaffold split deliberately induces distribution shift between training/calibration and test chemical types, this assumption may not strictly hold. Therefore, the nominal $1-\alpha$ coverage is not guaranteed by theory in this setting. Under the exchangeability assumption, regardless of the choice of $S(x)$, the conformal procedure ensures that the interval covers the true $y_{\text{test}}$ with probability $1-\alpha$ (marginally) \cite{angelopoulos2021gentle}. The crucial benefit is that if $S(x)$ captures heterogeneity in the error distribution, then $\hat q_{1-\alpha}S(x)$ will yield larger intervals where needed and smaller intervals elsewhere, improving efficiency while retaining marginal coverage under the stated assumption. Throughout the rest of the work, we denote the raw Mixture-of-Experts components as MoE$_E$, which is the epistemic component obtained via expert disagreement and MoE$_A$ which is the aleatoric component obtain via the per-expert variance. Their conformalized counterparts are written TESSERA$_{E}$ and TESSERA$_{A}$, respectively. 

\paragraph{Dataset and Preprocessing}
We train our model on the protein-ligand affinity dataset from \textsc{ChEMBL}31, yielding $350{,}400$ protein-ligand interaction (PLI) pairs with inhibition constants $K_i$ (nM). \textsc{ChEMBL} is a widely used, manually curated resource for bioactive molecules and bioactivities \cite{mendez2019chembl}. We transform affinities to $pK_i$ by $pK_i=-\log_{10}(K_i)$. We test out our model both on a random split (or independent and identically distributed split) and a scaffold-based split (or an out-of-distribution split), we utilize three random seeds ($\{0,11,42\}$) when using both strategies to train our models for statistical significance. The data is partitioned into  $60\%$ train, $20\%$ test, $10\%$ validation, and $10\%$ calibration. We obtain the results on the scaffold split to stress generalization across chemicals using the Bemis-Murcko split \cite{bemis1996properties}. 

\paragraph{Encoders}
The protein sequence is encoded using the smallest pretrained ESM-2 model \cite{lin2023evolutionary}. We apply parameter-efficient fine-tuning through low rank adaptation (LoRA) \cite{hu2022lora}. Small molecules are encoded using the SimSGT model, a masked-graph modeling encoder that uses a Simple GNN-based Tokenizer and remask decoding \cite{liu2023rethinking}. We first pretrain this model using the MolFormer data which consists of 111M PubChem compounds along with 1B compounds of the ZINC dataset \cite{ross2022large}. When training the model we completely tune the chemical encoder. We also evaluate the performance on alternate chemical encoders such as UniMol \cite{zhou2023unimol}, ChemBERTa-2 \cite{ahmad2022chemberta}, MolFormer \cite{ross2022large}, and ContextPred \cite{hu2019strategies}, these represent common choices in molecular representation tasks. Protein and chemical embeddings are concatenated and fed to a Mixture-of-Experts (MoE) fusion module with $K{=}4$ experts and a softmax gating network (simple linear gate). Each expert is a one-layer MLP that outputs a mean and a variance head. 

\paragraph{Baselines}
We compare against Monte Carlo Dropout \cite{gal2016dropout}, the GP-based RIO residual model \cite{qiu2019quantifying}, and eMOSAIC \cite{badkul2024emosaic}. For Monte Carlo Dropout, we keep the same backbone encoders as our method but remove the MoE fusion, replacing it with a simple concatenation of chem/protein embeddings followed by a small nonlinear MLP head. This avoids mixing MoE-style uncertainty with dropout so the comparison isolates the dropout heuristic. At inference, dropout remains active and we compute the predictive mean and variance from $T= 50$ stochastic forward passes. For RIO and eMOSAIC, we feed the pre-MoE concatenated embeddings into each method’s pipeline to ensure identical inputs across baselines. For eMOSAIC we set the number of clusters to $k=50$. All other hyperparameters and training details are reported in the appendix.

\paragraph{Evaluation Metrics}
We report root mean square error (RMSE), mean absolute error (MAE), pearson correlation, and spearman rank correlation for point accuracy, and negative log-likelihood (NLL) for probabilistic fit. 

The evaluation of the uncertainty centers on three main aspects: 
\begin{enumerate}
    \item Validity (Calibration) - Do we hit the promised coverage? 
    \item Efficiency (Sharpness) - How tight are intervals given validity?
    \item Adaptivity (Discriminative ability) - Do the interval widths track difficulty? 
\end{enumerate}

For validity, we evaluate the prediction interval coverage probability (PICP): Which is the empirical fraction of test samples whose true value $y_{\text{test}}$ falls inside the prediction interval. We expect this to be approximately $1-\alpha$ for a well-calibrated method. 

Efficiency quantifies how narrow the mean prediction intervals are, given that validity is satisfied. We measure this using the mean prediction interval width (MPIW) defined as the average width of the prediction intervals on the test set. In general, smaller MPIW indicates sharper and efficient predictions. However, MPIW must be interpreted together with validity, since, an extremely narrow interval is meaningless, if it fails to capture the true values. The goal is to minimize MPIW while maintaining nominal coverage. To enable fair comparison across different target ranges, we compute the Normalized Mean Prediction Interval Width (NMPIW) by scaling interval widths relative to the range of the true target values:

\begin{equation}
\text{NMPIW}=\frac{1}{n}\sum_{i=1}^{n}\frac{U_i-L_i}{y_{\max}-y_{\min}}   
\end{equation}

where  $U_i$ and  $L_i$ denote the upper and lower bounds of the prediction interval for the $i^{th}$ sample, $y_{max}$ and $y_{min}$ are the maximum and minimum observed target values, respectively. 

To effectively assess efficiency along with coverage, we utilize the coverage width-based criterion (CWC) \cite{khosravi2010lower, quan2014particle}, which penalizes under-coverage while favoring narrower intervals:

\begin{equation}
\text{CWC}=\text{NMPIW}\,\Big(1+\gamma(\text{PICP})\,e^{-\eta\big(\text{PICP}-\mu\big)}\Big)
\end{equation}

where

\begin{equation}
    \gamma(\text{PICP})=
\begin{cases}
0,& \text{PICP}\ge\mu\\
1,& \text{PICP}<\mu
\end{cases}
\end{equation}

and $\mu=1-\alpha$ denotes the nominal confidence level. 

If the coverage meets or exceeds the target ($\text{PICP}\ge\mu$), $\gamma=0$ the CWC reduces to NMPIW. In this case, efficiency comparisons depend solely on normalized interval width—smaller NMPIW values imply better sharpness. Over-coverage with tighter intervals is acceptable, whereas achieving nominal coverage with overly wide intervals indicates inefficiency. When coverage falls short, the exponential penalty term $\eta$ increases the CWC score proportionally to the shortfall. Typical choices for the penalty coefficient or $\eta$ range between $50$ and $100$ \cite{zhou2023real, khosravi2010lower, quan2014particle}. 

Lastly, for adaptivity, we use the following metrics:
\begin{enumerate}
    \item Area Under the Sparsification Error curve (AUSE): This metric comes from plotting a sparsification curve: we gradually remove a fraction of test points (from 0\% up to 100\%) in order of descending predicted uncertainty, and at each removal fraction we compute the model’s mean error on the remaining points. By removing the most uncertain points first, we should see the error drop faster if the uncertainty estimates are good. We compare this curve to an “oracle” curve obtained by removing points in order of actual error. The sparsification error is essentially the difference between the model curve and the oracle curve. AUSE is the area under this difference curve. Lower AUSE indicates our uncertainty ranking is close to optimal, we are almost as good as an oracle at identifying which points the model will likely get wrong. 
    \item Size-Stratified Coverage (SSC): To compute SSC, we will bin the test examples into groups based on their interval width. For each bin, we compute the coverage rate within that bin. This reveals whether our method is conditionally coverage-consistent – ideally, each bin should also have coverage near $1-\alpha$. A large deviation would indicate that our intervals are too small for some supposedly “easy” cases (a sign of misspecified $S(x)$). An adaptive method that truly works should exhibit no large coverage violations across interval sizes. 
\end{enumerate}

\section{Disentangling Uncertainty}\label{sec:disent}
\begin{figure}
    \centering
    \includegraphics[width=1\linewidth]{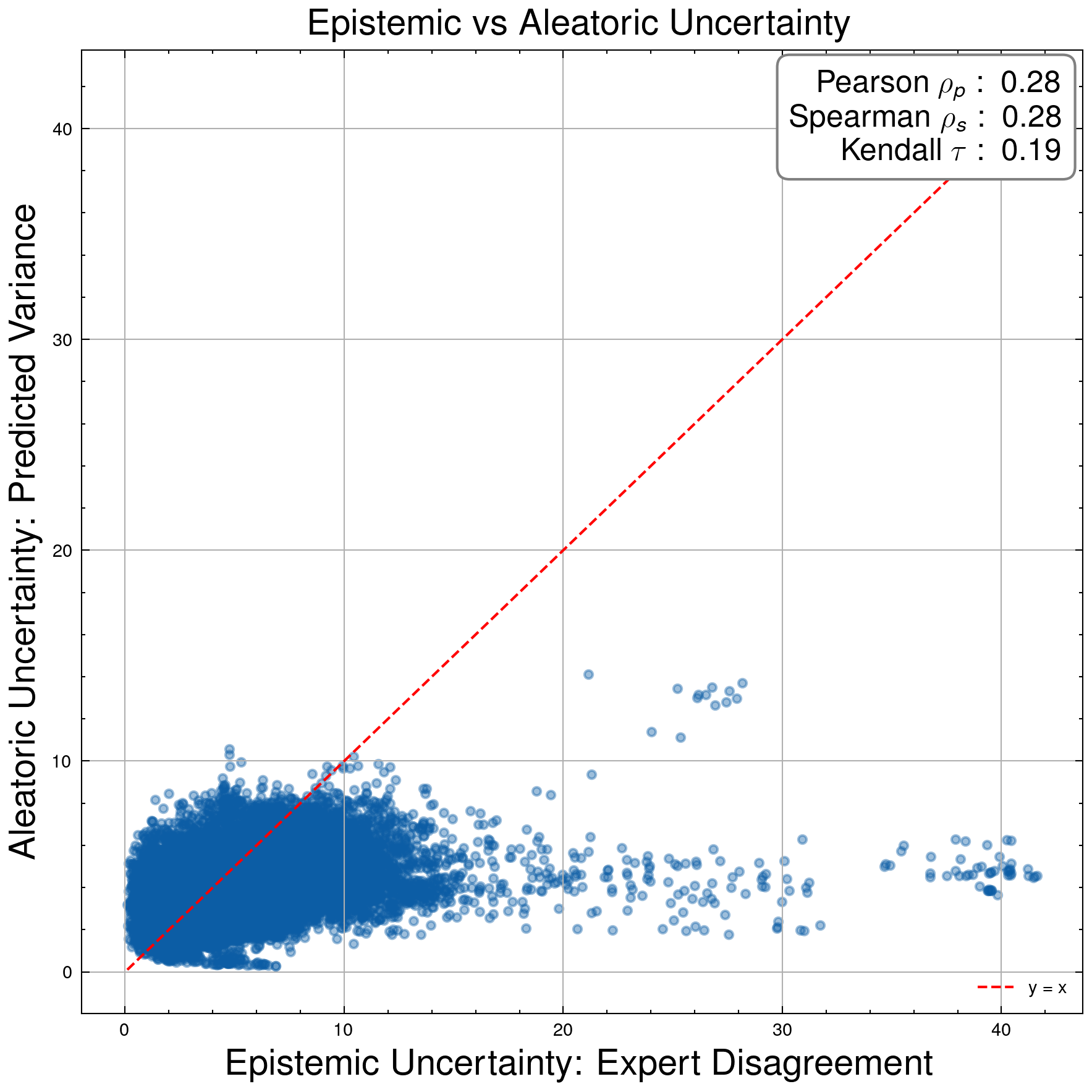}
    \caption{Disentangling Uncertainty }
    \label{fig:disent}
\end{figure}

We report two MoE-derived signals as operational representatives for the two uncertainty notions: MoE–E (expert disagreement) for epistemic/model uncertainty and MoE–A (per-expert predicted variance) for aleatoric/data uncertainty. In line with standard definitions, epistemic uncertainty reflects ignorance about the model and is reducible with data, whereas aleatoric uncertainty reflects irreducible data noise and can be heteroscedastic across inputs \cite{hullermeier2021aleatoric, kendall2017uncertaintiesneedbayesiandeep}. This conceptual split is widely used, but the precise decomposition is not identifiable from observations alone without additional assumptions, so empirical validation must be indirect. The scatter in Fig. A\ref{fig:disent} yields low to moderate correlation between MoE–E and MoE–A with Pearson correlation coefficient,  Spearman's rank correlation coefficient and Kendall rank correlation coefficient around 0.28, 0.28, and 0.19 respectively. This indicates the two signals are complementary and not redundant implying that they likely respond to different sources of difficulty. Rank-based coefficients corroborate the hypothesis that the ordering induced by one signal is only weakly aligned with the other, this is consistent with a partial disentanglement. Statistical tests such as Mann-Whitney U test and Welch's t test also return tiny p-values $1e-300$ and $1e-10$ respectively. The right interpretation is that the distributions differ, not that the effect is necessarily large. However, consistent with the literature’s cautions, we stop short of asserting a definitive causal decomposition. Instead, we use the signals operationally. They power the interval sizing and demonstrably improve validity, efficiency, and error-tracking under scaffold-OOD evaluation. We view the ablations above as future work to strengthen the evidence for the intended semantics. 

\section{Ablation}\label{sec:ablation}

\paragraph{OOD vs i.i.d Performance Comparison}
Across split strategies, TESSERA behaves similarly. On scaffold splits, both TESSERA$_A$, and TESSERA$_E$ reach near-nominal coverage, and on random split the coverage remains comparable, as seen in Table \ref{tab:uncertainty-split-comparison}. Interval widths are of the same order as well as the error-tracking signal. The combined CWC criterion shows small trade-offs, but the overall picture is consistent. With TESSERA, we are able to preserve the validity, efficiency, and adaptivity across both i.i.d. and OOD splits. 

\begin{table}[t]
\centering
\caption{\textbf{Comparison of uncertainty–quality metrics across splits.}
Test–set results for TESSERA variants under \textcolor{blue}{scaffold} and \textcolor{red}{random} splits with $\alpha{=}0.10$.
PICP targets $1-\alpha \approx 0.90$ (higher is better). MPIW and NMPIW assess interval efficiency (lower is better). AUSE evaluates uncertainty ranking quality (lower is better) and CWC is the combined criterion with $\eta{=}10$ (lower is better).
Best values are \textbf{bold}, and the second–best are \underline{underlined}.}
\label{tab:uncertainty-split-comparison}
\setlength{\tabcolsep}{6pt}
\small
\begin{tabular}{@{}clccccc@{}}
\toprule
\textbf{Split} & \textbf{Heuristic} &
PICP ($\uparrow\;{\approx}0.9$) &
MPIW ($\downarrow$) &
NMPIW ($\downarrow$) &
AUSE ($\downarrow$) &
CWC ($\downarrow$) \\
\midrule
\multirow{4}{*}{\textcolor{blue}{Scaffold}}
& MoE$_E$ & 0.480 & 1.49 & 0.060 & \underline{\textcolor{blue}{0.64}} & 4.12 \\
& MoE$_A$ & 0.400 & 1.07 & 0.040 & 0.64 & 6.98 \\
& TESSERA$_{E}$        & \textbf{\textcolor{blue}{0.910}} & 4.03 & 0.17 & 0.64 & \textbf{\textcolor{blue}{0.17}} \\
& TESSERA$_{A}$        & \underline{\textcolor{blue}{0.910}} & 4.76 & 0.20 & \textbf{\textcolor{blue}{0.58}} & \underline{\textcolor{blue}{0.20}} \\
\midrule
\multirow{4}{*}{\textcolor{red}{Random}}
& MoE$_E$ & 0.522 & 1.54 & 0.065 & \textbf{\textcolor{red}{0.512}} & \underline{\textcolor{red}{2.91}} \\
& MoE$_A$ & 0.417 & 1.06 & 0.045 & 0.624 & 5.59 \\
& TESSERA$_{E}$        & \underline{\textcolor{red}{0.899}} & 4.28 & 0.180 & \underline{\textcolor{red}{0.624}} & \underline{\textcolor{red}{0.36}} \\
& TESSERA$_{A}$        & \textbf{\textcolor{red}{0.902}} & 3.73 & 0.157 & 0.624 & \textbf{\textcolor{red}{0.16}} \\
\bottomrule
\end{tabular}
\end{table}

\paragraph{Comparison of different encoders}
Table \ref{tab:uncertainty-metrics-models} compares TESSERA across three very different molecule encoders: (i) ContextPred (GIN) \cite{hu2019strategies}, a graph-pretrained GNN that learns node/graph context signals, (ii) Uni-Mol \cite{zhou2023unimol}, a large 3D molecular foundation model, and (iii) SimSGT \cite{liu2023rethinking}, a recent masked-graph pretraining method with a Simple-GNN tokenizer and graph-transformer encoder. TESSERA$_A$ and TESSERA$_E$ achieve similar near-nominal coverage regardless of encoder being deployed, the PICP value ranges from 0.8 to 0.91, showing that the calibration step is model-agnostic and preserves reliability under both stronger (SimSGT) and weaker (or frozen Uni-Mol) feature backbones. Swapping encoders leaves the coverage guarantees of TESSERA intact and yields comparable efficiency and adaptivity. Better molecular representations translate into slightly tighter, that's the case for SimSGT encoder, but the method itself is encoder-agnostic. 
\begin{table}
\centering
\caption{\textbf{Comparison of uncertainty–quality metrics.} Test–set results for TESSERA variants across encoders. 
PICP targets $1-\alpha \approx 0.90$ (higher is better). MPIW and NMPIW assess interval efficiency (lower is better). 
AUSE evaluates uncertainty ranking (lower is better), and CWC is the combined criterion with $\eta{=}10$ and $\mu{=}0.9$ (lower is better).}
\label{tab:uncertainty-metrics-models}
\setlength{\tabcolsep}{6pt}
\small
\begin{tabular}{@{}clccccc@{}}
\toprule
Encoder & Heuristic & PICP ($\uparrow\;{\approx}0.9$) & MPIW ($\downarrow$) & NMPIW ($\downarrow$) & AUSE ($\downarrow$) & CWC ($\downarrow$) \\
\midrule
\multirow{4}{*}{ContextPred (GIN)}
  & TESSERA$_{E}$         & {0.88} & 4.62 & 0.20 & {0.71} & {0.43} \\
  & TESSERA$_{A}$         & {0.88} & 4.05 & 0.17 & 0.75 & {0.37} \\
  & MoE$_E$   & 0.31 & 1.49 & 0.06 & {0.71} & 23.09 \\
  & MoE$_A$   & 0.57 & 1.07 & 0.05 & 0.75 & 1.25 \\
\midrule
\multirow{4}{*}{UniMol}
  & TESSERA$_{E}$         & {0.89} & 4.62 & 0.19 &{0.67} & {0.41} \\
  & TESSERA$_{A}$         & {0.88} & 3.87 & 0.16 & {0.65} & {0.37} \\
  & MoE$_E$   & 0.37 & 1.22 & 0.05 & 0.67 & 10.08 \\
  & MoE$_A$   & 0.55 & 1.87 & 0.08 & 0.65 & 2.57 \\
\midrule
\multirow{4}{*}{SimSGT}
  & TESSERA$_{E}$         & \textbf{0.91} & 4.62 & 0.20 & \textbf{0.58} & \underline{0.20} \\
  & TESSERA$_{A}$         & \underline{0.91} & 4.05 & 0.17 & 0.64 & \textbf{0.17} \\
  & MoE$_E$   & 0.48 & 1.49 & 0.06 & \underline{0.58} & 4.16 \\
  & MoE$_A$   & 0.40 & 1.07 & 0.05 & 0.64 & 7.04 \\
\bottomrule
\end{tabular}
\end{table}

\begin{table}
\centering
\caption{\textbf{CWC values at different penalty coefficients $\eta$}. Comparison of coverage–width criterion (CWC) values for varying $\eta \in \{10, 50, 100\}$ using $\mu = 0.9$.}
\label{tab:cwc-eta}
\setlength{\tabcolsep}{6pt}
\small
\begin{tabular}{@{}lccc@{}}
\toprule
\textbf{Heuristic} & \textbf{CWC ($\eta{=}10$)} & \textbf{CWC ($\eta{=}50$)} & \textbf{CWC ($\eta{=}100$)} \\ 
\midrule
TESSERA$_{E}$        & \underline{0.20} & \underline{0.20} & \underline{0.20} \\
TESSERA$_{A}$        & \textbf{0.17}    & \textbf{0.17}    & \textbf{0.17}    \\
MoE$_E$  & 4.16             & 73.17            & 8498.93          \\
MoE$_A$  & 7.04             & 456.02           & 115440.97        \\
\bottomrule
\end{tabular}
\end{table}

\section{Hyperparameters}\label{sec:hyperparams}
\vspace{-3mm}
\begin{table}
\centering
\small
\caption{TESSERA hyperparameters with the SimSGT molecular encoder. Encoder-specific settings not listed (SimSGT, ESM2) use their original defaults.}
\begin{tabular}{ll}
\toprule
\textbf{Component} & \textbf{Setting} \\
\midrule
Conformal level $\alpha$        & 0.10 \\
GNN backbone                    & GIN \\
GNN dropout                     & 0.5 \\
Transformer encoder layers      & 4 \\
Learning rate                   & 1e-4 \\
Freeze GNN                      & False \\
Protein encoder                 & ESM-2 (8M) \\
LoRA (protein encoder)          & True, rank $r=8$, $\alpha=16$, dropout = 0.5 \\
LoRA layers                     & Key, Value, Query \\
Protein–ligand fusion module    & Mixture-of-Experts (MoE) \\
Gating network                  & MLP \\
Expert head architecture        & MLP \\
Number of experts               & 4 \\
\bottomrule
\end{tabular}
\label{tab:tessera-hyperparam}
\end{table}

\begin{table}[htbp]
\centering
\small
\caption{Hyperparameters for the UQ methods.}
\begin{tabular}{lll}
\toprule
\textbf{Method} & \textbf{Hyperparameter} & \textbf{Value} \\
\midrule
\multirow{1}{*}{TESSERA} 
  & Conformal level $\alpha$ & 0.10 \\
\midrule
\multirow{5}{*}{eMOSAIC} 
  & Kmeans Clusters k & 50 \\
  & MLP dimension & 128 \\
  & Dropout & 0.1 \\
  & Learning Rate & 1e-3 \\
  & Epochs & 50 \\
\midrule
\multirow{3}{*}{RIO-GP} 
  & Inducing Points & 50 \\
  & Iterations & 200 \\
  & Learning Rate & 1e-3 \\
\midrule
\multirow{5}{*}{MC Dropout} 
  & Dropout & 0.5 \\
  & Learning Rate & 1e-3 \\
  & Epochs & 50 \\
  & Stochastic $\tau$ passes & 50 \\
\bottomrule
\end{tabular}
\label{tab:baseline-hyperparam}
\end{table}

\begin{figure}
    \centering
    \includegraphics[width=1\linewidth]{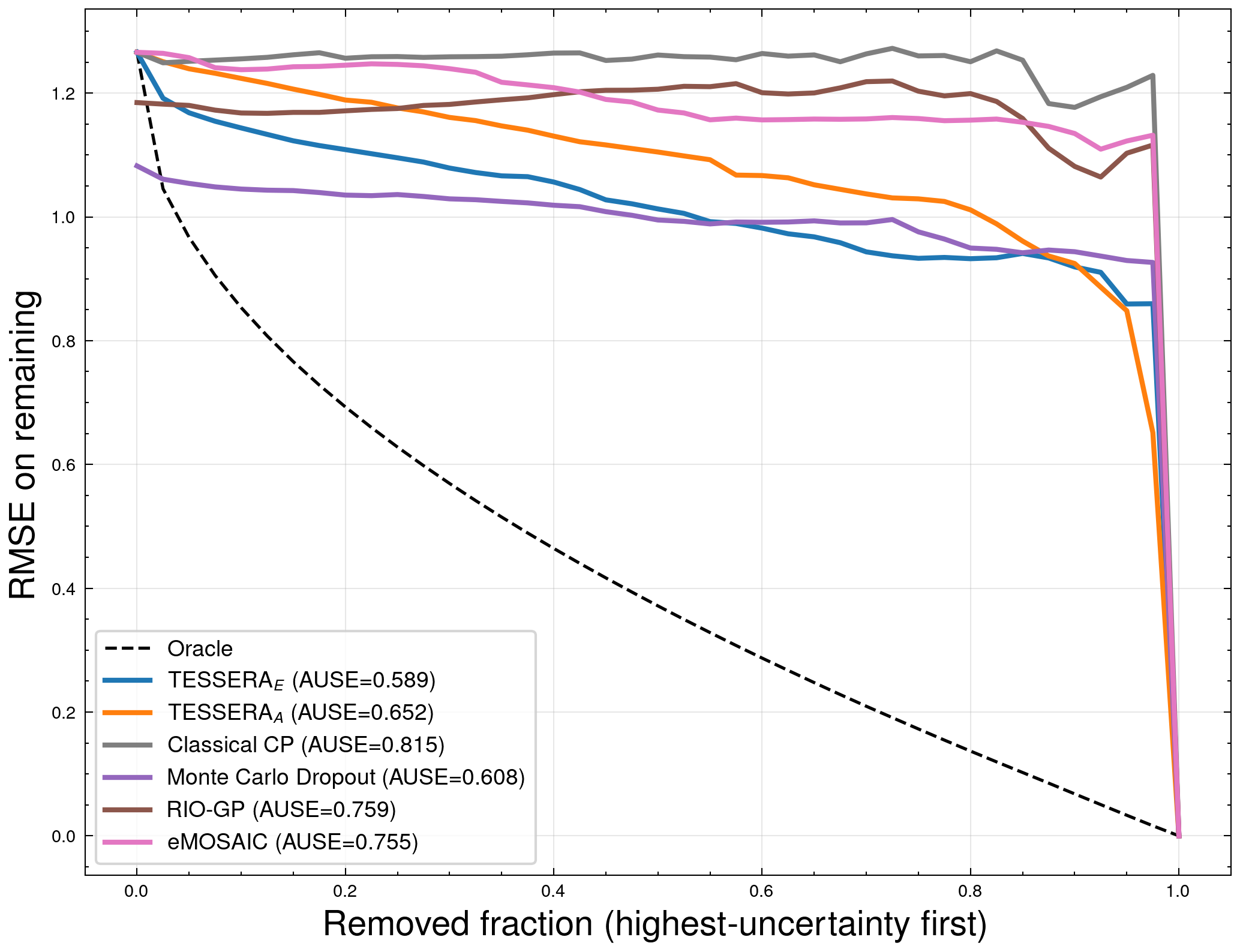}
    \caption{Risk–coverage (sparsification) curves: RMSE on the remaining data after removing the fraction on the x-axis with the highest predicted uncertainty. The dashed black curve is the oracle (removing points by true error). The legend reports AUSE (area between each curve and the oracle, lower is better).}
    \label{fig:ause_plot}
\end{figure}

\end{document}